\documentclass{article} 
\usepackage[preprint]{colm2025_conference}

\usepackage{microtype}
\usepackage{hyperref}
\usepackage{url}
\usepackage{booktabs}
\usepackage{caption} 
\usepackage{graphicx}
\usepackage{tcolorbox}
\usepackage{amsmath}
\usepackage{algorithm}
\usepackage{algorithmic}
\usepackage{multirow}
\usepackage{wrapfig}
\usepackage{enumitem}
\usepackage{fancyvrb}
\usepackage{fvextra} 
\usepackage{lineno}
\usepackage{xcolor}
\usepackage{subcaption} 

\usepackage{amsmath,amsfonts,bm}

\def\eqref#1{equation~\ref{#1}}

\def\1{\bm{1}}

\def\rvc{{\mathbf{c}}}

\def\rvp{{\mathbf{p}}}

\def\rvs{{\mathbf{s}}}

\def\rvy{{\mathbf{y}}}
\def\rvz{{\mathbf{z}}}

\def\vs{{\bm{s}}}

\DeclareMathAlphabet{\mathsfit}{\encodingdefault}{\sfdefault}{m}{sl}
\SetMathAlphabet{\mathsfit}{bold}{\encodingdefault}{\sfdefault}{bx}{n}

\def\gD{{\mathcal{D}}}

\def\gU{{\mathcal{U}}}

\newcommand{\E}{\mathbb{E}}

\definecolor{darkblue}{rgb}{0, 0, 0.5}
\hypersetup{colorlinks=true, citecolor=darkblue, linkcolor=darkblue, urlcolor=darkblue}

\newcommand{\think}{\texttt{<T>}}
\newcommand{\thinkend}{\texttt{</T>}}

\newcommand{\method}{{{\textsc{Trelawney}}}}

\title{Looking beyond the next token}

\author{
  \hspace{2.5em}
  {Abitha Thankaraj\thanks{Equal contribution. Correspondence to \texttt{athankar@cs.cmu.edu}} }
  \hspace{2em}
  {\textbf{Yiding Jiang}\footnotemark[1]} 
  \hspace{2em}
  \textbf{J. Zico Kolter} 
  \hspace{2em}
  \textbf{Yonatan Bisk} \\[5pt]
  \hspace{0.35\linewidth}\textnormal{Carnegie Mellon University}
}

\begin{document}

\ifcolmsubmission
\linenumbers
\fi

\maketitle

\begin{abstract}
The structure of causal language model training assumes that each token can be accurately predicted from the previous context.
This contrasts with humans' natural writing and reasoning process, where goals are typically known before the exact argument or phrasings. While this mismatch has been well studied in the literature, the working assumption has been that architectural changes are needed to address this mismatch. We argue that rearranging and processing the training data sequences can allow models to more accurately imitate the true data-generating process, and does not require any other changes to the architecture or training infrastructure. We demonstrate that this technique \method{} and the inference algorithms derived from it allow us to improve performance on several key benchmarks that span planning, algorithmic reasoning, and story generation tasks. Finally, our method naturally enables the generation of long-term goals at no additional cost. We investigate how using the model’s goal-generation capability can further improve planning and reasoning. Additionally, we believe \method{} could potentially open doors to new capabilities beyond the current language modeling paradigm.

\end{abstract}

\section{Introduction}

Next-token prediction (NTP) is the primary objective for training sequence models.
This objective involves a technique called \emph{teacher forcing}~\citep{williams1989learning}, where the model's predicted output at each step is replaced with the ground truth from the real dataset.
One of teacher forcing's benefits is that it accelerates the training by providing the model with the correct previous output, so the learning does not suffer from error accumulation, and the gradient update is more stable.
Another crucial benefit is that it enables parallelism and hardware acceleration in training because the model can simultaneously process all time steps, rather than sequentially waiting for its own predictions.
However, \citet{bachmann2024pitfalls} argue that models trained with teacher forcing often fail to learn long-range dependencies, latching onto local patterns and surface-level correlations instead.

Several recent methods have been proposed to alleviate the issues of teacher forcing. 
One popular approach is \emph{multi-token prediction}, where the model learns to predict multiple tokens at the same time~\citep{bachmann2024pitfalls, gloeckle2024better, liu2024deepseek}. 
Another family of approaches involves modifying the training objective to predict both the next token for a prefix and the previous token for a suffix by modifying the model architecture~\citep{hu2025the}.
Most of these approaches either involve nontrivial modification to the model architecture or make the learning process much harder by forcing the model to predict multiple tokens at the same time.

\begin{figure}[t]
  \centering
  \includegraphics[trim=0.2cm 11.5cm 0cm 2cm,clip,width=\textwidth]{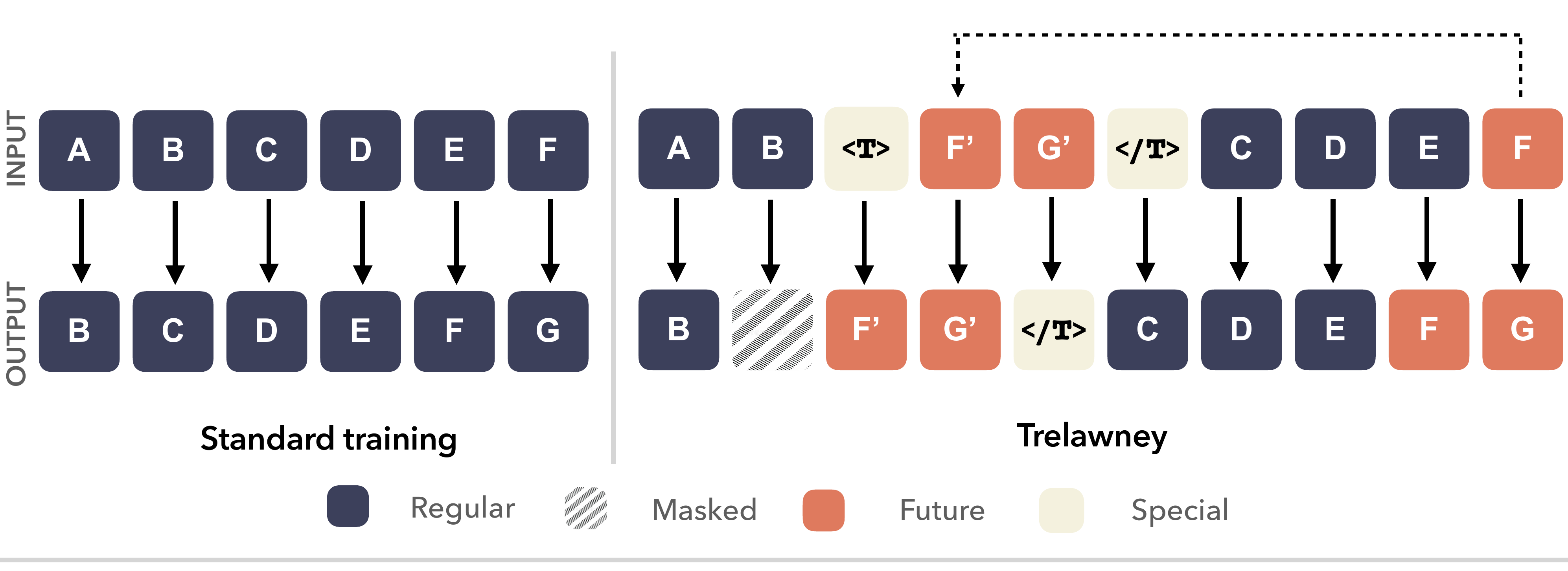}
  \vskip 0.8mm
  \includegraphics[trim=0.5cm 14cm 0cm 4cm,clip,width=\textwidth]{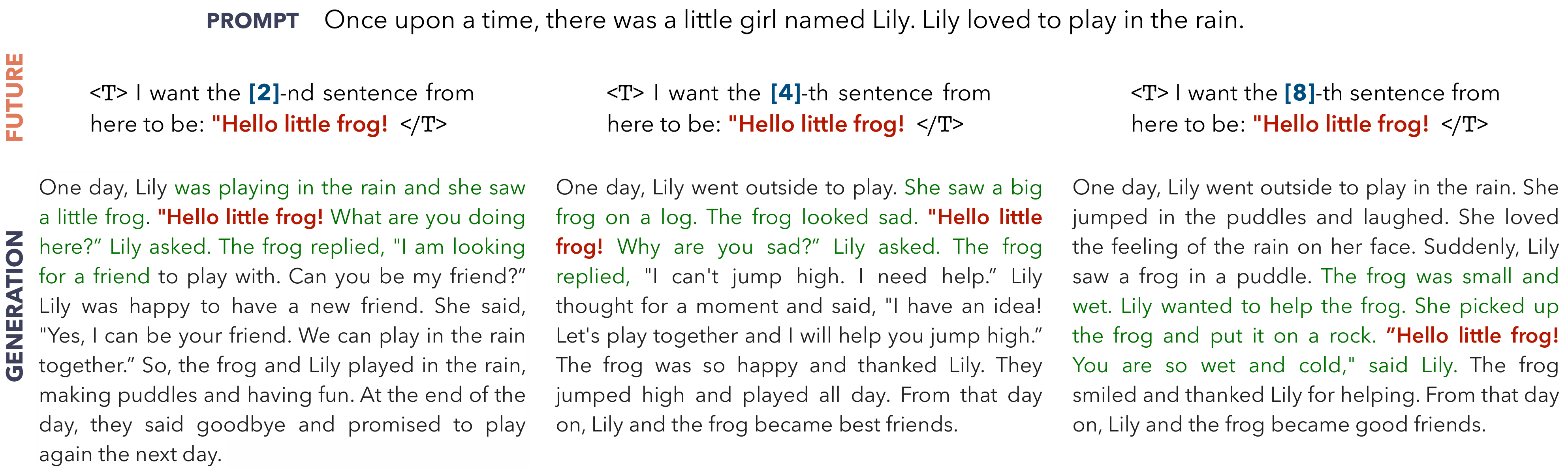}
  \caption{\method. \textbf{(Top)} We modify the training data by inserting tokens from the future, delimited with special tokens $\think$ and $\thinkend$, into the modified sequences. By first predicting tokens in the future, the model is encouraged to learn the tokens pertaining to what it will generate in the future (i.e., \texttt{F'G'}), and the path leading to the future (i.e., \texttt{CDE}) as well as the actual future (i.e., \texttt{FG}) will be easier to predict. \textbf{(Bottom)} Illustration of \method's effect during generation. The top is the prompt, and the middle is different future tokens. The generations are coherent and read naturally.}
  \label{fig:method_fig}
\vspace{-1em}
\end{figure}

In this work, we investigate a data-centric approach to address these limitations. In contrast to the strictly sequential nature of traditional training, the flow of information in real-world tasks is highly non-linear. Instead of modifying the model architecture, our method \method{} modifies the training data by introducing alternative factorizations that embed inductive biases directly. Concretely, we augment the training corpus by interleaving it with special lookahead tokens --- \texttt{<T>} and \texttt{</T>} --- that encapsulate future information (see Figure~\ref{fig:method_fig}). The exact placement and content of these tokens can be determined either randomly or with task-specific knowledge.
We hypothesize that this augmentation makes learning the long-term dependencies easier and imbues the model with the capacity to plan ahead.
Furthermore, these modified training data naturally teach the model to guide the generation towards the future information, so the lookahead tokens can also let users exert fine-grained control over the long-term generation. 

This work contributes a simple data-rearrangement procedure, \method{}\footnote{The name is inspired by the seer who predicts the future in the Harry Potter series.}, which results in both improved task performance in domains otherwise difficult for models trained with next token prediction, by decoupling the discussion of the training objective from the underlying data generating function the model needs to learn. Results are presented on planning and algorithmic reasoning tasks, in addition to conditional story generation.
\vspace{-1em}

\section{Preliminaries}
\label{preliminaries}
Consider a sequence of tokens 
\(\mathbf{y} = (y_1, y_2, \ldots, y_T)\), where each token \(y_t\) belongs to a fixed vocabulary \(V\). Suppose that $\mathbf{y}$ follows a distribution $P(\mathbf{y})$,
an auto-regressive model \(P_{\theta}\) factorizes the joint probability of 
\(\mathbf{y} = (y_1, y_2 ... y_T)\) as follows:
$ \label{eq:1}
    P_\theta(\mathbf{y}) \;=\; \prod_{t=1}^{T} p_\theta\!\bigl(y_t \mid \mathbf{y}_{<t}
    \bigr)
$
\! where \(\mathbf{y}_{<t} = (y_1, \ldots, y_{t-1})\) denotes all tokens before index \(t\).

\textbf{Next token prediction.} We train the model parameters \(\theta\) to maximize the likelihood of each 
token under the ground truth context (teacher forcing) from the training corpus. Concretely, if our training data consists of 
\(N\) sequences \(\{\mathbf{y}^{(i)}\}_{i=1}^N\), each of length \(T\), then the 
maximum likelihood estimate of \(\theta\) is obtained by:
$
\theta^* \;=\; \arg \max_\theta 
\sum_{i=1}^N \sum_{t=1}^{T} 
\log \, p_\theta\!\left(y_t^{(i)} \mid \mathbf{y}_{<t}^{(i)}\right).
$

\textbf{Inference.} At inference time, the model predicts the next token by either sampling or selecting the most likely token, conditioned on an optional context $\mathbf{c}$ (e.g., question). In the greedy 
setting, the next token \(\widehat{y}_t\) is
$
\hat{y}_t \;=\; \arg \max_{y_t}\; 
p_\theta\!\left(y_t \mid \widehat{\mathbf{y}}_{<t}, \,\mathbf{c}\right),
$
where, \(\widehat{y}_{<t}\) denotes the model's own generated tokens up to position \(t\).
\vspace{-0.8em}
\subsection{Pitfalls of teacher forcing}
\label{subsec:pitfall}
\citet{bachmann2024pitfalls} highlighted that teacher forcing can cause subtle failure modes in language models.
In particular, they identified: 

\textit{Clever Hans Cheat.} When training with teacher-forcing, the model is provided with ground truth prefixes (e.g., $v_{\text{start}}, v_1, \dots, v_{i-1}$) that include parts of the answer. This extra information can enable the model to ``cheat'' by simply copying the easy tokens that follow without learning the true underlying plan. 
    
\textit{Indecipherable Token Problem.} 
Because the later tokens can be easily predicted using the Clever Hans cheat, the crucial early decision receives insufficient gradient signal. 
This early token becomes ``indecipherable'' since its correct prediction relies on long-range planning that is effectively bypassed during teacher-forced training.

\textit{Exposure bias.} During inference, the model would likely make a mistake because the model has not learned the indecipherable token.
The mismatch between training (where the model always sees the correct previous tokens) and inference (where it must rely on its own predictions) can lead to a cascading sequence of errors.
\vspace{-1em}

\section{\method{}}

Standard decoder-only transformers typically learn the factorization of a sequence defined in Equation~\ref{eq:1}. We introduce a data augmentation scheme that modifies the given sequence $\mathbf{y}$ as follows:
first select a point $d$ and insert a sequence of $k$ tokens, $\rvz = (z_1, z_2, \dots, z_k)$, delimited with special tokens $\think$ and $\thinkend$. Concretely, we have the following augmentation:
\begin{align*}
    (y_1 \  y_2\  \dots \  y_T) \Longrightarrow  (y_1 \  y_2 \ \dots \  y_{d} \  \think \  \rvz \ \thinkend \   y_{d+1} \ \dots\  y_{T-1} \  y_T)
\end{align*}
The choices of $d$, $k$, and the content of $\mathbf{z}$ are flexible, and we present several strategies. %
\vspace{-0.8em}
\subsection{Augmentation schemas}
\noindent\textbf{Copying.} We can directly copy a part of the sequence from a point after $y_d$ to between the special tokens. For $s$ such that $d\! < \! s \! \leq \! T\! -\!k$, $\rvz$ is the subsequence $\rvy_{s:s+k}$, resulting in
\begin{align*}
    \Tilde{\rvy}_\text{copy} &\equiv  y_1 \  y_2\  \dots y_{d}  
    \  \texttt{<T>} \  \rvy_{s:s+k} \ \texttt{</T>}\   y_{d+1}\  \dots \  y_{T-1} \  y_T
\end{align*}
The choice of $\rvz$ can have a significant impact on the behavior of the resulting model. 
For example, for certain types of data, there are \emph{decision points} where there are many different possible futures. 
These points are good candidates for choosing $d$.
Analogously, we can choose $\rvz$ to be \emph{future tokens} that indicate which future is being generated.
These terms are intentionally defined loosely since the design space is large.(\S\ref{choice_of_z_star_graph}, \S \ref{choice_of_z_scc})
Without this prior information, we can also select $d$, $s$, and $k$ randomly, which can be useful for generic language modeling. %

\noindent\textbf{Positional information.} 
In the previous approach, $d$ and $s$ can vary between different data points.
This can be problematic if two sequences have very different values of $s - d$. 
Intuitively, this makes the modeling task harder because there may be conflicting information between different sequences. 
For example, suppose $\rvy^{1}$ and $\rvy^2$ share the same prefixes, $\rvy^1_{:d} = \rvy^2_{:d}$ but the relevant future tokens are at locations with large differences. %

To mitigate this conflict, we introduce additional \emph{positional information} into the future tokens, $\zeta(k, \rvz)$. For example, we can have:
\begin{align*}
    \zeta(k, \rvz) &= ``\texttt{I want the \texttt{[k]}th sentence from here to be } \mathbf{z}'', \\
    \Tilde{\rvy}_{\text{copy+pos}} &\equiv y_1 \ y_2 \ \dots \ y_d \ \think \ \zeta(k, \rvz) \ \thinkend \ y_{d+1} \ \dots \ y_{d+k} \ \dots \ y_n 
\end{align*}
Once again, the exact design of the positional information can be problem-dependent (\S~\ref{choice_of_zeta}). The position provided does not need to be highly accurate as long as it reduces potential conflict.
Similarly, the copied text $\rvz$ can be a copy of a sequence from the future, $\rvy_{d:d+k}$, but does not need to be identical, so long as it contains relevant information (e.g., paraphrase).

We express $\zeta$ in natural language because this allows the model to integrate $\zeta$ with its pretraining knowledge and also lets the user specify different goals.
Alternatively, this can also be done with special tokens.
It is also possible to introduce various agentic behaviors via $\zeta$, though we only conduct a preliminary investigation in this direction.
\vspace{-.8em}
\subsection{Dataset Construction and Training Objective}

\noindent\textbf{Dataset construction.} Our goal is to introduce additional capabilities via the augmentation schema shown above. 
However, it is desirable to do so without hurting the traditional language modeling ability of the model.
To accomplish this, we train on both regular text and augmented text simultaneously.
Specifically, given an original dataset \(D = \{\rvy^{(i)}\}_{i=1}^N\) and an augmentation schema $\texttt{aug}$.
We can construct a distribution for the original dataset, and a distribution for the augmented dataset:
\begin{align*}
    \gD(\rvs) = \frac{1}{N}\sum_{i=1}^N \mathbb{I}\left\{\rvs = \rvy^{(i)}\right\}, \quad \gD_\texttt{aug}(\rvs) = \frac{1}{N}\sum_{i=1}^N \mathbb{I}\left\{\rvs = \texttt{aug}(\rvy^{(i)})\right\}.
\end{align*}
For a probability $p$ that controls how much of the training distribution comprises the original data, the training distribution is the following mixture:
$
    \gD'(\vs) = p \, \gD(\rvs) + (1-p) \, \gD_\texttt{aug}(\rvs).
$

\noindent\textbf{Training and loss function.} During training, the model parameters are optimized using a standard cross-entropy loss with teacher forcing on $\gD'$.
This allows us to take advantage of all existing engineering optimizations for training language models.
One caveat for training with the new dataset $\gD'$ is that choosing the decision point and future tokens arbitrarily will result in a large portion of sequences with the next token being \texttt{</T>} at arbitrary locations.

This would distract from the learning process and does not help learning the underlying distribution, since the special tokens are synthetically introduced.
Instead, we modify the regular cross-entropy loss by masking the special start token, \texttt{<T>}:
\[
\mathcal{L}(\gD') = - \E_{\rvy \sim \gD'}\left[\frac{1}{|\rvy|}\sum_{j=1}^{|\rvy|} \mathbb{I}\{y_j \neq \texttt{<T>}\} \log P(y_j \mid \rvy_{<j})\right].
\] \label{loss_fn}
\!Here, \(\mathbb{I}\{y_j \neq \think\}\) ensures no loss is computed for the prediction of the special token \think. Note that we do not exclude the loss on $\thinkend$ because there is a utility to predicting the closing of the future tokens, which we will elaborate on below.
\vspace{-0.8em}
\subsection{Inference}

\noindent\textbf{Standard autoregressive generation.} The model generates sequences autoregressively without any intervention, following any standard decoding algorithm.

\noindent\textbf{\think-generation.} We aim to enable the model to explicitly consider future context at appropriate decision points, to improve its ability to plan ahead. At each decision point $y_d$ in sequence generation, we explicitly insert the special token \think. Subsequently, \textbf{(a)} either the model generates the sequence $\mathbf{z}$ autonomously, enabling it to create plausible future plans, or \textbf{(b)} incorporates a user-specified sequence $\mathbf{z}$, enhancing controllability. 
Recall that during the training process, we compute the loss on the $\thinkend$ token, this allows the model to generate future goals, which can then be used for conditional generation.
In contrast to existing methods such as \citet{hu2025the} that require specific decoding mechanisms, our approach can use any off-the-shelf decoding algorithm.

\vspace{-0.8em}

\section{Experiments}
\vspace{-1em}
We hypothesize that prioritizing predictions of challenging tokens or strategic goals can improve generation quality. To test this, we conduct experiments across three synthetic and natural language tasks.
Synthetic tasks serve as a controlled environment where specific token-level or semantic unit challenges --- otherwise difficult to disentangle in natural language --- can be precisely studied. 

We begin with two synthetic benchmarks: the star graph (\S~\ref{subsec: star_graph}), a didactic example used to highlight limitations of next-token prediction even in simple tasks, and an algorithmic reasoning benchmark (\S~\ref{subsec: clrs_text}), which allow us to test whether explicitly adding future goals or anchor points improves performance in simple reasoning tasks. Finally, we extend our analysis to story generation (\S~\ref{subsec: tiny_stories}) to assess controllable generation and planning.

Our experiments are designed to answer the following questions:
\vspace{-10pt}
\begin{itemize}[noitemsep,leftmargin=20pt]
\item Does \method{} enable the generation of long-term goals? 
\item Does \method{} training improve planning when using autoregressive inference?
\item Do the goals generated explicitly by the model improve planning?
\item Does explicitly providing goal sequences $\mathbf{z}$'s improve controllability?
\end{itemize}
\vspace{-1em}
\subsection{Star Graph}
\label{subsec: star_graph}

\begin{figure}[t]
  \centering
  \includegraphics[trim=14cm 10.8cm 12cm 8.5cm,clip,width=0.8\textwidth]{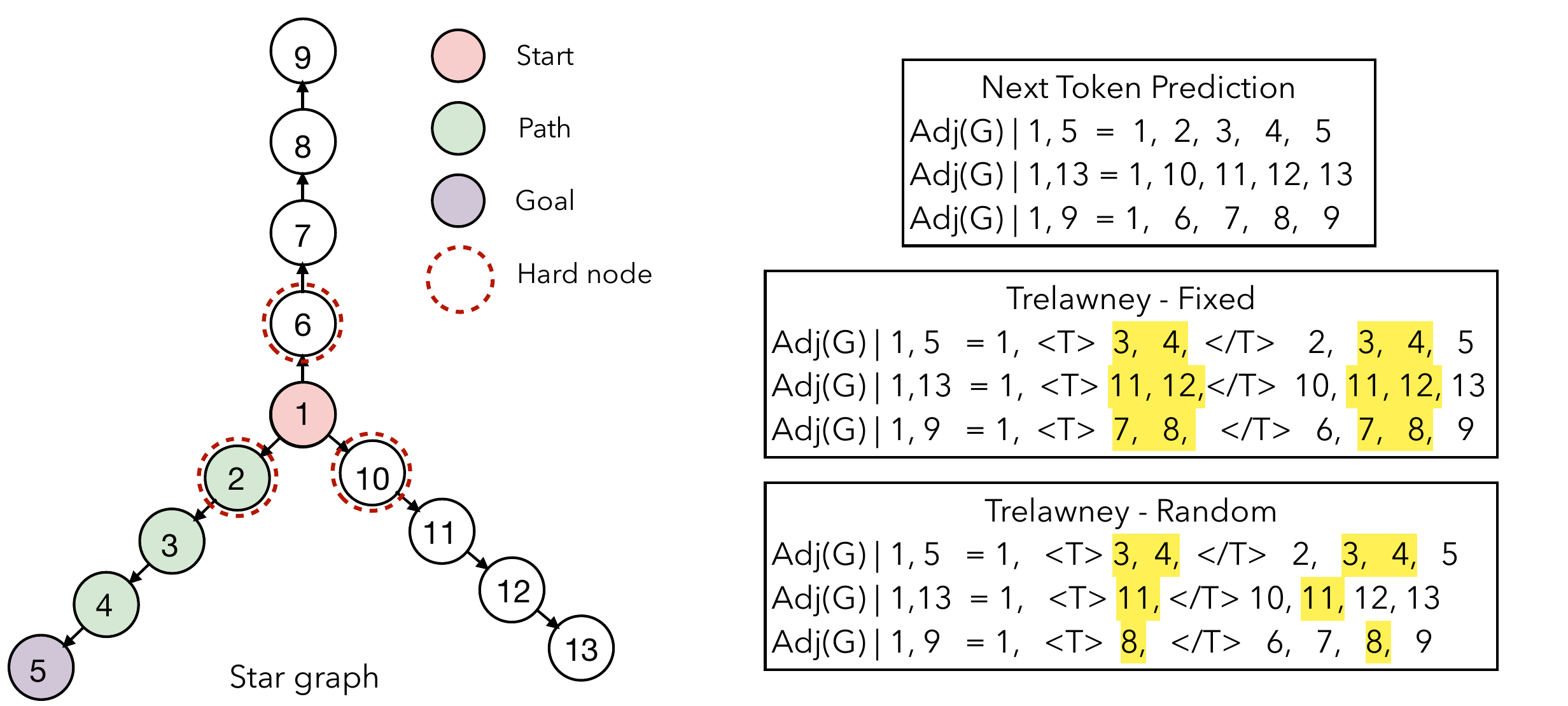}
  \caption{In the star graph, there are key ``hard nodes" that indicate the moment of branching, after which the path and goal become clear.  Above is a visualization of the construction of $\gD'$ for the star graph with corresponding linearizations.} \label{fig:star_graph}
  \vspace{-1em}
\end{figure}

The star graph is a simple path-finding problem introduced by \citet{bachmann2024pitfalls}, where, given a directed graph $G(d,n)$ with degree $d$ and path length $n$, the objective is to find a path from the start node to the goal node (Figure~\ref{fig:star_graph}). Despite its simplicity, traditional next-token prediction (NTP) struggles on this task. A key challenge is that the critical decision point occurs at $v_1$, the first node after $v_{\text{start}}$. This node is hard to predict because $v_{\text{start}}$ has many outgoing edges. As discussed in Section~\ref{subsec:pitfall}, teacher forcing can lead to undesirable behavior on this simple dataset.

\noindent\textbf{Dataset and Augmentation Schema.} To mitigate these issues, we introduce a future subgoal $\rvz$, as any contiguous subsection of the path in $[v_2, v_\text{goal})$. This modification compels the model to generate a meaningful intermediate plan rather than simply copying the full ground truth prefix. As a result, the model receives a stronger learning signal for critical early decision-making. Each example $\rvy = (\rvp,  \rvc)$ in the dataset is a prefix and completion pair. The prefix $\textbf{p}$ is given by the adjacency list of G followed by the $v_{start}, v_{goal} =$. The completion $\textbf{c}$ is the path $v_\text{start}, v_1, v_2, ... v_\text{goal}$, i.e., $\rvp \equiv \text{Adj}(G) \mid v_{\text{start}}, v_{\text{goal}} = $ and $\rvc \equiv v_{\text{start}}, v_1, v_2, \dots, v_{\text{goal}}$.

Our augmentation schema $\rvy \Longrightarrow \Tilde{\rvy}_\text{copy} $ for this task is as follows: 
\begin{align*}
\rvy &\equiv \text{Adj}(G) \mid v_{\text{start}}, v_{\text{goal}} = v_{\text{start}}, v_1, v_2, \dots, v_{\text{goal}} \\
\Tilde{\rvy}_{\text{copy}} &\equiv \text{Adj}(G) \mid v_{\text{start}}, v_{\text{goal}} = v_{\text{start}}, \, \think \  \rvz, \thinkend \   v_1, \dots, v_{\text{goal}}
\end{align*}

 \paragraph{Choice of $\rvz$.}\label{choice_of_z_star_graph} We vary $\rvz$ (a contiguous subsequence of future tokens) across experiments and ablations (see Figure~\ref{fig:star_graph}). Its role is to guide planning by indicating a subgoal on the path from $v_\text{start}$ to $v_\text{goal}$. We exclude $v_1$ to avoid the Clever Hans cheat discussed above. We also exclude $v_\text{goal}$ so that the model learns the long-term dependency between start and goal without having direct access to the goal token. An ablation study confirms that including $v_\text{goal}$ does not yield further improvements.

\noindent\textbf{Training.} Data for all experiments are generated programmatically using the official implementation\footnote{\url{https://github.com/gregorbachmann/Next-Token-Failures}}. 
Although we use pretrained models, each node remains a single token in the tokenizer. All models are trained on $200,000$ examples as described in Appendix~\ref{appendix:training_details}. 
We follow standard teacher forcing training with two variants of augmentation schemas.

\emph{\method{}-fixed:} In a single training run, the choice of $\rvz$ is fixed across all examples. Specifically, $\rvz$ is chosen as a contiguous sequence of 1 to 4 nodes with a fixed start and end point across all sequences in the dataset (Figure~\ref{fig:star_graph}).

\emph{\method{}-random:} $\rvz$ can vary between examples. We randomly select any contiguous subsequence of the path after $v_1$ to serve as $\rvz$ in $\Tilde{\rvy}_\text{copy}$. We do not include $v_1$ (the hard node) as part of $\rvz$ (Figure~\ref{fig:star_graph}). Without fixed positional information, the model learns to generate its own goals of varying lengths. We observe that this variant is successful in solving longer planning problems.

\noindent\textbf{Evaluation.} We evaluate the models on 5,000 held-out examples for each graph, reporting the accuracy of the generated path compared to the ground truth. For the next-token prediction baseline, we evaluate the model using standard autoregressive generation. For models trained with \method{}, we assess both standard autoregressive and \think-generation. In the conditional setting, the model uses either model-generated $\rvz$’s as goals or user-provided ground truth ``future goals'' as hints.
Standard autoregressive generation allows us to test whether \method{} improves regular generation. \think-generation demonstrates whether the model has learned to generate plausible future goals and use these goals for better planning. By providing intermediate hints, we evaluate if the model can leverage these cues to solve the larger planning problems.

\noindent\textbf{Results.} On shorter graphs $G(2,5)$, $G(5,5)$, $G(10,5)$, $G(20,5)$, training with \method{} improves autoregressive generation at no additional cost, suggesting that the model implicitly learns to plan better (possibly due to pre-caching or breadcrumbs proposed by \citet{wulanguage}) and can generate long-term goals. For longer graphs $G(2,10)$, the \method-random variant can complete the task when the model is used to generate its own subgoal sequence $\mathbf{z}$, indicating that model-generated goals can improve planning. \method-random is notably more performant on graphs with longer paths when compared to \method-fixed, We hypothesize that the random variant's flexibility in future goal selection has a regularizing effect that allows the model to learn better representations, but leave further investigation to future work. Both variants of \method{} succeed when user-provided goal sequences are provided, showing that explicit goal hints allow for better controllability.(See: Table \ref{tab:accuracies}) Further, ablations conducted on larger models (See \ref{appendix:ablations_model_capacity}) show that the ability to plan for future tokens improves with increase in model capacity.

\vspace{-1em}
\subsection{Algorithmic Reasoning}
\label{subsec: clrs_text} 
\vspace{-1em}
CLRS-Text~\citep{markeeva2024clrs}\footnote{\url{https://huggingface.com/datasets/tomg-group-umd/CLRS-Text-train}} is a %
benchmark of algorithmic reasoning. The input is the algorithm name, followed by a step-by-step reasoning trace and the final answer. We pick a representative example from algorithms that require backtracking, i.e., tasks that benefit from information of future states. We choose strongly-connected-components, a step-by-step sequential prediction task where each step is longer than one token, and report results on it. The trace contains the execution of Tarjan's algorithm \citep{tarjan1972depth}, which computes strongly connected components in linear time by performing a depth-first search that tracks low-link values and uses a stack to detect cycles.

\begin{table}[t]
\centering
\small
\begin{tabular}{@{}l@{\hspace{3pt}}l@{\hspace{0pt}}c@{\hspace{5pt}}c@{\hspace{5pt}}c@{\hspace{5pt}}c
                    @{\hspace{1em}}l@{\hspace{3pt}}c@{\hspace{5pt}}c@{\hspace{5pt}}c@{\hspace{5pt}}c@{\hspace{5pt}}c@{}}
\toprule
       & & \multicolumn{4}{@{}c@{\hspace{1em}}}{\textbf{Path planning} G(*,*)} 
       & & \multicolumn{5}{c}{\textbf{Alg Reasoning} scc-} \\
       
                            & &  G(2,5) & G(5,5) & G(20,5) & G(2,10) &                       & scc-4 & scc-5 & scc-11 & scc-12 & scc-15 \\
\midrule
\multirow{4}{*}{\rotatebox{90}{\footnotesize AutoReg.}} &
      NTP                       & \underline{0.50} & \underline{0.20} & \underline{0.05} & 0.50 &                     & \textbf{1.00} & 0.99 & 0.62 & 0.57 & 0.27 \\
      & \method{}               &      &      &      &      &                     &      &      &      &      &      \\
      & \hspace{1em} -- Fixed   & \textbf{1.00} & \textbf{1.00} & \textbf{1.00} & 0.52 &       -- Rule-Based & \textbf{1.00} & \textbf{1.00} &  0.73 & 0.62 & 0.31 \\
      & \hspace{1em} -- Random  & \textbf{1.00} & \textbf{1.00} & \textbf{1.00} & 0.50 &       -- Random     &   \textbf{1.00}   &  0.978    &  0.718    &   0.706   &   0.476    \\

\midrule
\multirow{4}{*}{\rotatebox{90}{\footnotesize Generated}} &
      NTP                       &  --  &  --  &  --  &  --  &                     & --   & --   & --   & --   &  --  \\
      & \method{}               &      &      &      &      &                     &      &      &      &      &      \\
      & \hspace{1em} -- Fixed   & \textbf{1.00} & \textbf{1.00} & \textbf{1.00} & 0.57 &       -- Rule-Based & \textbf{1.00} & \textbf{1.00} & 0.73 & 0.65 & 0.34 \\
      & \hspace{1em} -- Random  & \textbf{1.00} & \textbf{1.00} & \textbf{1.00} & \underline{0.91} &       -- Random     &   \textbf{1.00}   &   \underline{0.998}   &   0.776   &  \underline{0.79}    &   \underline{0.512}  \\

\midrule
\multirow{4}{*}{\rotatebox{90}{\footnotesize Specified}} &
      NTP                       &  --  & --   & --   &  --  &                     &  --  & --   & --   &  --  &  --   \\
      & \method{}               &      &      &      &      &                     &      &      &      &      &      \\
      & \hspace{1em} -- Fixed   & \textbf{1.00} & \textbf{1.00} & \textbf{1.00} & \textbf{1.00} &       -- Rule-Based & \textbf{1.00} & \textbf{1.00} & \textbf{0.84} & 0.76 & 0.47 \\
      & \hspace{1em} -- Random  & \textbf{1.00} & \textbf{1.00} & \textbf{1.00} & \underline{0.91} &       -- Random     &   \textbf{1.00}   &   \underline{0.998}   &   \underline{0.828}   &  \textbf{0.812}    &   \textbf{0.544}   \\

\bottomrule
\end{tabular}
\caption{\method{} outperforms next token prediction across synthetic domains:\\ Path planning (\textit{Star Graph}) and Algorithmic reasoning (\textit{Strongly connected components}). \\
Autoreg, Generated and Specified refer to inference methods of standard autoregressive generation, model generated \texttt{<T>}-generation and user specified \texttt{<T>}-generation respectively.\\
Fixed, Random and Rule-Based indicate the data augmentation strategies used in training. \\
\textbf{Bold} indicates the best and \underline{underline} indicates the second best.\vspace{-1em}}
\label{tab:accuracies}
\end{table}
\noindent\textbf{Dataset and Augmentation Schema.} In each example $\rvy = (\rvp,  \rvc)$ of the strongly-connected-components subset, the prefix $\textbf{p}$ is given by the adjacency matrix of the initial graph. The completion $\textbf{c}$ is graph execution traces
of the algorithm followed by the final answer, i.e., $\rvp \equiv \text{Adj}(G) = $ and $\rvc \equiv t_1, t_2 \dots t_n | F $ where $t_i$ is the state of the graphical trace and $F$ is the final answer.
Our augmentation schema $\rvy \Longrightarrow \Tilde{\rvy}_\text{copy} $ for this task is as follows:
\begin{align*}
\rvy &\equiv \text{algo: }  \text{Adj}(G) = t_1, t_2, \dots, t_n | F \\
\Tilde{\rvy}_\text{copy} &\equiv \text{algo: }  \text{Adj}(G) = t_1,  \think \   \rvz \  \thinkend\   t_2, \dots, t_n | F
\end{align*}
Unlike the star graph task --- where failure typically occurs at a single critical decision point --- the algorithmic reasoning tasks involve multiple branching points where errors can accumulate. In the strongly connected components subset, the state sequence $t$ represents the graph execution trace and comprises multiple tokens, each corresponding to a distinct graph state. By segmenting the trace into these meaningful units, our augmentation schema is better able to capture intermediate reasoning steps and guide the model’s planning process throughout the entire execution trace.

\noindent \textbf{Choice of $\rvz$.}\label{choice_of_z_scc} For simplicity, we fix the decision point $y_d$ at the second state in each trace. Unlike the star graph task, where the hard tokens are easy to extract, algorithmic reasoning tasks do not present a clear failure point --- there can be many points in the trace at which misprediction causes the entire generation to diverge. We only pick $\rvz$ as a complete step $t_i$ in the trace and how $i$ is determined for each variant.

\noindent \textbf{Training.} Data for all experiments are sub-selected from the original dataset. We train a single model on problems of varying sizes. Since we do not test for length generalization, we only report accuracies on problem sizes present in the training corpus. For strongly connected components, all models are trained on 60,000 examples. We train two variants (See \ref{example:scc}) of our method:

\emph{\method-rule-based}: For every example in $\gD_\text{aug}$, $\rvz$ is chosen as the first change in the trace provided. The position of $\rvz$ in the trace varies across graph sizes and graphs.

\emph{\method-random}: $\rvz$ is chosen as a single random state in the trace provided. 

\noindent \textbf{Evaluation.} We evaluate on 500 examples (\begin{small}{CLRS-Text-test}\end{small}) %
per graph length. We evaluate the models similar to the star-graph setting, and report the accuracies of the final answer. 

\noindent \textbf{Results} Results from Figure~\ref{fig:accuracies_scc} show a trend that \method-Random consistently improves on next token prediction when using \think-generation and, surprisingly, in standard autoregressive generation as well. \method-Rule-Based although being chosen more strategically, performs worse than \think-Random.

\subsection{Natural Language Planning}
\label{subsec: tiny_stories}
Story generation is inherently a planning task, as it requires the model to set long-term goals, maintain consistency, and control narrative flow. Although many narratives include complex elements like detailed backstories and conflict resolution, even simpler story generation requires planning to produce coherent and logically structured output.
We evaluate our approach on the \textbf{Tiny Stories} dataset from \citet{eldan2023tinystories}. In this task, the aim is to generate coherent stories conditioned on specified goals. This benchmark tests whether our strategy of inserting future tokens can enhance planning in natural language generation.

\noindent\textbf{Dataset and Augmentation Schema.} Each example $\rvy = (\rvp, \rvc)$ is a prefix–completion pair extracted from a story. We use a sentence parser to segment each story into individual sentences or phrases. If a story is split into sentences $s_1, s_2, \dots, s_n$, the prefix $\rvp$ is the beginning of the story (e.g., $s_1\ s_2$) and the completion $\rvc$ is the remainder (i.e., $s_3\ s_4\ \dots\ s_n$).

Our augmentation schema $\rvy \Longrightarrow \Tilde{\rvy}_\text{copy+pos} $ is defined as: 
\begin{align*}
\rvy &\equiv s_1 \ s_2 \ \dots \ s_n \\
\Tilde{\rvy}_{\text{copy+pos}} &\equiv s_1 \ s_2 \ \dots \ s_d \ \think \ \zeta(k, s_{d+k}) \ \thinkend \ s_{d+1} \ \dots \ s_{d+k} \ \dots \ s_n \\
\zeta(k, s) &= ``\texttt{I want the [k]-th sentence from here to be [s] }'', 
\end{align*}
\noindent\textbf{Choice of $\zeta(k, s)$.}\label{choice_of_zeta} We choose decision points randomly at the end of the $k$-th sentence in the document, as the position to to insert $\zeta(k, s)$.  The subgoal \(\texttt{[s]}\) is defined in $\zeta(d, s)$ as extracted from the corresponding sentence $s_{d+k}$. 

\noindent\textbf{Training.} All models are trained on $300,000$ examples from the Tiny Stories dataset for 1 epoch using the masked cross-entropy loss specified in \S~\ref{loss_fn} (See App.~\ref{appendix:training_details}). 

\vspace{-4pt}
\begin{itemize}[noitemsep,leftmargin=20pt]
    \item \emph{Next-token prediction:} We follow standard teacher forced training on the dataset $\gD$.
    \item \emph{\method-implicit-pos:} We insert goals into $\gD_{\text{aug}}$ by specifying how far ahead the goal is in natural language, without labeling each sentence with a numeric step index. 
    This probes if the model can discern positional information without explicit labels.
    \item \emph{\method-explicit-pos:} In addition to indicating how far ahead the goal is, we label each sentence with explicit stepwise indices after inserting the special tokens in $\gD_{\text{aug}}$. This gives the model more positional information about the goal.
\end{itemize}

\textbf{Evaluation}: We follow the evaluation protocol used by \citet{hu2025the} and use GPT-4 as a judge to rate 100 generated stories from each model. The stories are anonymized and shuffled to prevent any information leakage about the author before being passed to the judge. Each evaluation is repeated over 6 trials. We report the win rate with binomial confidence intervals computed at a 95\% significance level.
 
\textbf{RQ1: Does \method{} improve goal reaching ability i.e., resulting in more controllable generation?}
We compare the completions from few-shot prompts on the baseline with those obtained by explicitly specifying goals on \method-implicit. Qualitatively, we observe that models trained with \method{} generate stories that more effectively reach the intended long-term goals (see Figure~\ref{fig:method_fig}).
Quantitatively, we observe that GPT-4 prefers \method{} to few-shot prompts on next-token-prediction, $76.53\%$ of the time, with a confidence interval of [72.9\%, 79.9\%]. This result suggests that \method is much more effective at controllable generation than few-shot prompting. We provide more details and ablations of few few-shot prompting variants used in \ref{appendix:gpt4_judge}. 

\textbf{RQ2: Preference on stories generated by standard autoregressive generation.}
We compare the standard autoregressive generations from models trained with \method{} and models trained with NTP.
Quantitatively, we observe that GPT-4 prefers autoregressive generations on \method to next-token-prediction, $40.35\%$ of the time, with a binomial confidence interval of [44.5\%, 36.2\%]. The justification for judgements appear to be preferences in ending of the stories, which qualitatively, does not appear to affect factors such as coherence and creativity. We provide examples of GPT-4 preference evaluations in \ref{appendix:gpt4_judge}. 

\noindent\textbf{Ablations.} Adding explicit positions as in \method--explicit provides better fine grained control on the goal position than with no explicit labels by sentence in \method--implicit. 

\begin{figure}[t]
  \centering
  \begin{subfigure}[b]{0.49\textwidth }
    \centering
      \includegraphics[trim=0.2cm .1cm .5cm 0.1cm,clip,width=\textwidth]{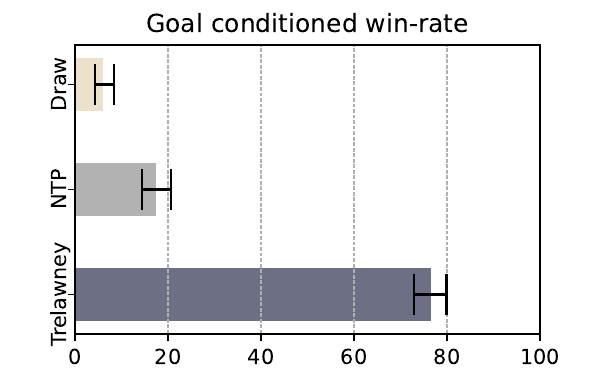}
    \label{fig:goal_conditioned}

  \end{subfigure}
  \hfill
  \begin{subfigure}[b]{0.49\textwidth}
    \centering
  \includegraphics[trim=0.2cm .1cm .5cm 0.2cm,clip,width=\textwidth]{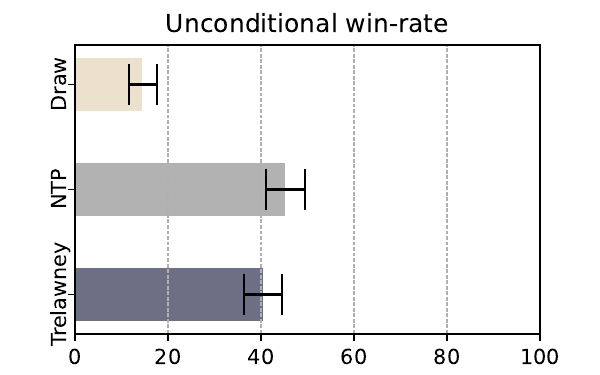}
    \label{fig:win_rate_unconditional}
  \end{subfigure}
  \caption{Our evaluation of story generation demonstrates greatly improved performance when goal-conditioned, without hurting the performance of unconditional generation.}
  \label{fig:win_rate_combined}
  \vspace{-1.5em}
\end{figure}

We evaluate perplexity on Wikitext to verify that \method{} maintains language model performance, with results comparable to the baseline.We provide more details in \ref{appendix:perplexity}
\vspace{-1em}

\section{Related work}
\vspace{-0.8em}
\noindent \textbf{Next token prediction.} 
\citet{bachmann2024pitfalls} characterizes two failure that occur in next-token prediction, those that emerge from (1) teacher-forced training, and (2) those emerging at inference, where errors compound (the so-called snowballing failure). Much of the prior work~\citep{arora2022exposure, ross2011reduction} has focused primarily on the inference-time errors. In contrast, during training, the maximum likelihood estimation (MLE) objective treats all tokens equally. However, \citet{bigelow2024forking} provides empirical evidence that tokens contribute unequally to the overall performance, suggesting that some tokens are inherently more critical than others. Relatedly, \citet{lin2024rho} propose leveraging a stronger model to identify and prioritize these important tokens for more efficient pretraining. \citet{nye2021show} introduce scratchpads which augment the model’s input with intermediate reasoning steps to support multi-step problem solving. \citet{goyal2023think} introduces pause tokens at training and inference, as a mechanism for delayed next-token prediction, which improves performance across several language tasks.

\noindent \textbf{Architectural changes.} \cite{du2023autoregressive, yin2024semformer} propose architectural modifications to improve planning. \citet{qi2020prophetnet} predict multiple n-grams rather than a single token at a time.
Similarly, \citet{gloeckle2024better, liu2024deepseek} incorporate these insights and have empirically validated the approach at scale. \citet{hu2025the} introduce an additional encoder that learns representations for both the next and previous states simultaneously. \citet{frydenlund2024mystery} explores alternate encoder-based architectures for planning.

\noindent \textbf{Controllable generation.} Our work is also related to the literature of controllable generation, where the models are conditioned to follow high-level goals or guidelines provided through explicit instructions or auxiliary inputs.
Prominent methods include 
\citet{keskar2019ctrl, dathathri2019plug, krause2020gedi}.
More recent models can be controlled via \emph{prompting}~\citep{brown2020language, wei2022chain}. In comparison, \method{} does not require a curated dataset or additional classifiers and achieves fine-grained temporal control.

\noindent \textbf{Non-causal sequence modeling.} offers an alternative to the traditional autoregressive, left-to-right generation constraint by allowing the model to use both past and future context~\citep{gu2017non, gong2022diffuseq,nolte2024transformers}. \citet{bavarian2022efficient} propose a ``fill in the middle'' strategy which changes the data ordering, while T5 \citep{raffel2020exploring} incorporates span corruption, \(\sigma\)-GPT \citep{pannatier2024sigma} uses on-the-fly order modulation, MLM-$\gU$ \citep{kitouni2024factorization} uses uniform masking similar to the diffusion objective and XLNet \citep{yang2019xlnet} leverages permutation-based training. 
Inference-time strategies, such as tree generation~\citep{welleck2019non}, have also been explored. Beyond language modeling, video prediction~\citep{han2019video, vondrick2016anticipating} similarly relies on non-causal prediction of future frames or states. In control tasks and world modeling~\citep{lecun2022path, hafner2023mastering, linlearning}, non-causal approaches provide a more comprehensive representation of environmental dynamics, thereby enhancing long-term planning.

\vspace{-1em}
\section{Discussion}
\vspace{-0.5em}
We introduce a method to improve controllable generation. Our small-scale finetuning experiments are effective but limited to synthetic or constrained scenarios. Our preliminary results suggest that the approach holds promise for adaptation to more general settings. We leave extending to  more generalizable settings such as pretraining to future work. Beyond simple copying behaviors, our method opens the door to using reinforcement learning to control generation based on the information enclosed by the special tokens. One remaining challenge is determining when the model should leverage these capabilities; uncertainty metrics may offer a promising solution.

\section*{Acknowledgements}
We thank Sean Welleck, Yingshan Chang, and Amanda Bertsch for discussions on early scoping of this work, Daman Arora and Fahim Tajwar for inputs on experimental details and Vaishnavh Nagarajan and Nishant Subramani for feedback on the project. AT acknowledges funding from Google. YJ is supported by the Google PhD Fellowship.

\bibliography{bibliography}

\begin{thebibliography}{38}
\providecommand{\natexlab}[1]{#1}
\providecommand{\url}[1]{\texttt{#1}}
\expandafter\ifx\csname urlstyle\endcsname\relax
  \providecommand{\doi}[1]{doi: #1}\else
  \providecommand{\doi}{doi: \begingroup \urlstyle{rm}\Url}\fi

\bibitem[Arora et~al.(2022)Arora, Asri, Bahuleyan, and Cheung]{arora2022exposure}
Kushal Arora, Layla~El Asri, Hareesh Bahuleyan, and Jackie Chi~Kit Cheung.
\newblock Why exposure bias matters: An imitation learning perspective of error accumulation in language generation.
\newblock \emph{arXiv preprint arXiv:2204.01171}, 2022.

\bibitem[Bachmann \& Nagarajan(2024)Bachmann and Nagarajan]{bachmann2024pitfalls}
Gregor Bachmann and Vaishnavh Nagarajan.
\newblock The pitfalls of next-token prediction.
\newblock \emph{arXiv preprint arXiv:2403.06963}, 2024.

\bibitem[Bavarian et~al.(2022)Bavarian, Jun, Tezak, Schulman, McLeavey, Tworek, and Chen]{bavarian2022efficient}
Mohammad Bavarian, Heewoo Jun, Nikolas Tezak, John Schulman, Christine McLeavey, Jerry Tworek, and Mark Chen.
\newblock Efficient training of language models to fill in the middle.
\newblock \emph{arXiv preprint arXiv:2207.14255}, 2022.

\bibitem[Bigelow et~al.(2024)Bigelow, Holtzman, Tanaka, and Ullman]{bigelow2024forking}
Eric Bigelow, Ari Holtzman, Hidenori Tanaka, and Tomer Ullman.
\newblock Forking paths in neural text generation.
\newblock \emph{arXiv preprint arXiv:2412.07961}, 2024.

\bibitem[Brown et~al.(2020)Brown, Mann, Ryder, Subbiah, Kaplan, Dhariwal, Neelakantan, Shyam, Sastry, Askell, et~al.]{brown2020language}
Tom Brown, Benjamin Mann, Nick Ryder, Melanie Subbiah, Jared~D Kaplan, Prafulla Dhariwal, Arvind Neelakantan, Pranav Shyam, Girish Sastry, Amanda Askell, et~al.
\newblock Language models are few-shot learners.
\newblock \emph{Advances in neural information processing systems}, 33:\penalty0 1877--1901, 2020.

\bibitem[Dathathri et~al.(2019)Dathathri, Madotto, Lan, Hung, Frank, Molino, Yosinski, and Liu]{dathathri2019plug}
Sumanth Dathathri, Andrea Madotto, Janice Lan, Jane Hung, Eric Frank, Piero Molino, Jason Yosinski, and Rosanne Liu.
\newblock Plug and play language models: A simple approach to controlled text generation.
\newblock \emph{arXiv preprint arXiv:1912.02164}, 2019.

\bibitem[Deepseek et~al.(2024)Deepseek, Liu, Feng, Xue, Wang, Wu, Lu, Zhao, Deng, Zhang, Ruan, et~al.]{liu2024deepseek}
Deepseek, Aixin Liu, Bei Feng, Bing Xue, Bingxuan Wang, Bochao Wu, Chengda Lu, Chenggang Zhao, Chengqi Deng, Chenyu Zhang, Chong Ruan, et~al.
\newblock Deepseek-v3 technical report.
\newblock \emph{arXiv preprint arXiv:2412.19437}, 2024.

\bibitem[Du et~al.(2023)Du, Mei, and Eisner]{du2023autoregressive}
Li~Du, Hongyuan Mei, and Jason Eisner.
\newblock Autoregressive modeling with lookahead attention.
\newblock \emph{arXiv preprint arXiv:2305.12272}, 2023.

\bibitem[Eldan \& Li(2023)Eldan and Li]{eldan2023tinystories}
Ronen Eldan and Yuanzhi Li.
\newblock Tinystories: How small can language models be and still speak coherent english?
\newblock \emph{arXiv preprint arXiv:2305.07759}, 2023.

\bibitem[Frydenlund(2024)]{frydenlund2024mystery}
Arvid Frydenlund.
\newblock The mystery of the pathological path-star task for language models.
\newblock \emph{arXiv preprint arXiv:2410.13779}, 2024.

\bibitem[Gloeckle et~al.(2024)Gloeckle, Idrissi, Rozi{\`e}re, Lopez-Paz, and Synnaeve]{gloeckle2024better}
Fabian Gloeckle, Badr~Youbi Idrissi, Baptiste Rozi{\`e}re, David Lopez-Paz, and Gabriel Synnaeve.
\newblock Better \& faster large language models via multi-token prediction.
\newblock \emph{arXiv preprint arXiv:2404.19737}, 2024.

\bibitem[Gong et~al.(2022)Gong, Li, Feng, Wu, and Kong]{gong2022diffuseq}
Shansan Gong, Mukai Li, Jiangtao Feng, Zhiyong Wu, and LingPeng Kong.
\newblock Diffuseq: Sequence to sequence text generation with diffusion models.
\newblock \emph{arXiv preprint arXiv:2210.08933}, 2022.

\bibitem[Goyal et~al.(2023)Goyal, Ji, Rawat, Menon, Kumar, and Nagarajan]{goyal2023think}
Sachin Goyal, Ziwei Ji, Ankit~Singh Rawat, Aditya~Krishna Menon, Sanjiv Kumar, and Vaishnavh Nagarajan.
\newblock Think before you speak: Training language models with pause tokens.
\newblock \emph{arXiv preprint arXiv:2310.02226}, 2023.

\bibitem[Gu et~al.(2017)Gu, Bradbury, Xiong, Li, and Socher]{gu2017non}
Jiatao Gu, James Bradbury, Caiming Xiong, Victor~OK Li, and Richard Socher.
\newblock Non-autoregressive neural machine translation.
\newblock \emph{arXiv preprint arXiv:1711.02281}, 2017.

\bibitem[Hafner et~al.(2023)Hafner, Pasukonis, Ba, and Lillicrap]{hafner2023mastering}
Danijar Hafner, Jurgis Pasukonis, Jimmy Ba, and Timothy Lillicrap.
\newblock Mastering diverse domains through world models.
\newblock \emph{arXiv preprint arXiv:2301.04104}, 2023.

\bibitem[Han et~al.(2019)Han, Xie, and Zisserman]{han2019video}
Tengda Han, Weidi Xie, and Andrew Zisserman.
\newblock Video representation learning by dense predictive coding.
\newblock In \emph{Proceedings of the IEEE/CVF international conference on computer vision workshops}, pp.\  0--0, 2019.

\bibitem[Hu et~al.(2025)Hu, Ahn, Liu, Xu, Tomar, Langford, Jayaraman, Lamb, and Langford]{hu2025the}
Edward~S. Hu, Kwangjun Ahn, Qinghua Liu, Haoran Xu, Manan Tomar, Ada Langford, Dinesh Jayaraman, Alex Lamb, and John Langford.
\newblock The belief state transformer.
\newblock In \emph{The Thirteenth International Conference on Learning Representations}, 2025.
\newblock URL \url{https://openreview.net/forum?id=ThRMTCgpvo}.

\bibitem[Keskar et~al.(2019)Keskar, McCann, Varshney, Xiong, and Socher]{keskar2019ctrl}
Nitish~Shirish Keskar, Bryan McCann, Lav~R Varshney, Caiming Xiong, and Richard Socher.
\newblock Ctrl: A conditional transformer language model for controllable generation.
\newblock \emph{arXiv preprint arXiv:1909.05858}, 2019.

\bibitem[Kitouni et~al.(2024)Kitouni, Nolte, Williams, Rabbat, Bouchacourt, and Ibrahim]{kitouni2024factorization}
Ouail Kitouni, Niklas~S Nolte, Adina Williams, Michael Rabbat, Diane Bouchacourt, and Mark Ibrahim.
\newblock The factorization curse: Which tokens you predict underlie the reversal curse and more.
\newblock \emph{Advances in Neural Information Processing Systems}, 37:\penalty0 112329--112355, 2024.

\bibitem[Krause et~al.(2020)Krause, Gotmare, McCann, Keskar, Joty, Socher, and Rajani]{krause2020gedi}
Ben Krause, Akhilesh~Deepak Gotmare, Bryan McCann, Nitish~Shirish Keskar, Shafiq Joty, Richard Socher, and Nazneen~Fatema Rajani.
\newblock Gedi: Generative discriminator guided sequence generation.
\newblock \emph{arXiv preprint arXiv:2009.06367}, 2020.

\bibitem[LeCun(2022)]{lecun2022path}
Yann LeCun.
\newblock A path towards autonomous machine intelligence version 0.9. 2, 2022-06-27.
\newblock \emph{Open Review}, 62\penalty0 (1):\penalty0 1--62, 2022.

\bibitem[Lin et~al.()Lin, Du, Watkins, Hafner, Abbeel, Klein, and Dragan]{linlearning}
Jessy Lin, Yuqing Du, Olivia Watkins, Danijar Hafner, Pieter Abbeel, Dan Klein, and Anca Dragan.
\newblock Learning to model the world with language.

\bibitem[Lin et~al.(2024)Lin, Gou, Gong, Liu, Shen, Xu, Lin, Yang, Jiao, Duan, et~al.]{lin2024rho}
Zhenghao Lin, Zhibin Gou, Yeyun Gong, Xiao Liu, Yelong Shen, Ruochen Xu, Chen Lin, Yujiu Yang, Jian Jiao, Nan Duan, et~al.
\newblock Rho-1: Not all tokens are what you need.
\newblock \emph{arXiv preprint arXiv:2404.07965}, 2024.

\bibitem[Markeeva et~al.(2024)Markeeva, McLeish, Ibarz, Bounsi, Kozlova, Vitvitskyi, Blundell, Goldstein, Schwarzschild, and Veli{\v{c}}kovi{\'c}]{markeeva2024clrs}
Larisa Markeeva, Sean McLeish, Borja Ibarz, Wilfried Bounsi, Olga Kozlova, Alex Vitvitskyi, Charles Blundell, Tom Goldstein, Avi Schwarzschild, and Petar Veli{\v{c}}kovi{\'c}.
\newblock The clrs-text algorithmic reasoning language benchmark.
\newblock \emph{arXiv preprint arXiv:2406.04229}, 2024.

\bibitem[Nolte et~al.(2024)Nolte, Kitouni, Williams, Rabbat, and Ibrahim]{nolte2024transformers}
Niklas Nolte, Ouail Kitouni, Adina Williams, Mike Rabbat, and Mark Ibrahim.
\newblock Transformers can navigate mazes with multi-step prediction.
\newblock \emph{arXiv preprint arXiv:2412.05117}, 2024.

\bibitem[Nye et~al.(2021)Nye, Andreassen, Gur-Ari, Michalewski, Austin, Bieber, Dohan, Lewkowycz, Bosma, Luan, et~al.]{nye2021show}
Maxwell Nye, Anders~Johan Andreassen, Guy Gur-Ari, Henryk Michalewski, Jacob Austin, David Bieber, David Dohan, Aitor Lewkowycz, Maarten Bosma, David Luan, et~al.
\newblock Show your work: Scratchpads for intermediate computation with language models.
\newblock 2021.

\bibitem[Pannatier et~al.(2024)Pannatier, Courdier, and Fleuret]{pannatier2024sigma}
Arnaud Pannatier, Evann Courdier, and Fran{\c{c}}ois Fleuret.
\newblock $\sigma$-gpts: A new approach to autoregressive models.
\newblock In \emph{Joint European Conference on Machine Learning and Knowledge Discovery in Databases}, pp.\  143--159. Springer, 2024.

\bibitem[Qi et~al.(2020)Qi, Yan, Gong, Liu, Duan, Chen, Zhang, and Zhou]{qi2020prophetnet}
Weizhen Qi, Yu~Yan, Yeyun Gong, Dayiheng Liu, Nan Duan, Jiusheng Chen, Ruofei Zhang, and Ming Zhou.
\newblock Prophetnet: Predicting future n-gram for sequence-to-sequence pre-training.
\newblock \emph{arXiv preprint arXiv:2001.04063}, 2020.

\bibitem[Raffel et~al.(2020)Raffel, Shazeer, Roberts, Lee, Narang, Matena, Zhou, Li, and Liu]{raffel2020exploring}
Colin Raffel, Noam Shazeer, Adam Roberts, Katherine Lee, Sharan Narang, Michael Matena, Yanqi Zhou, Wei Li, and Peter~J Liu.
\newblock Exploring the limits of transfer learning with a unified text-to-text transformer.
\newblock \emph{Journal of machine learning research}, 21\penalty0 (140):\penalty0 1--67, 2020.

\bibitem[Ross et~al.(2011)Ross, Gordon, and Bagnell]{ross2011reduction}
St{\'e}phane Ross, Geoffrey Gordon, and Drew Bagnell.
\newblock A reduction of imitation learning and structured prediction to no-regret online learning.
\newblock In \emph{Proceedings of the fourteenth international conference on artificial intelligence and statistics}, pp.\  627--635. JMLR Workshop and Conference Proceedings, 2011.

\bibitem[Tarjan(1972)]{tarjan1972depth}
Robert Tarjan.
\newblock Depth-first search and linear graph algorithms.
\newblock \emph{SIAM journal on computing}, 1\penalty0 (2):\penalty0 146--160, 1972.

\bibitem[Vondrick et~al.(2016)Vondrick, Pirsiavash, and Torralba]{vondrick2016anticipating}
Carl Vondrick, Hamed Pirsiavash, and Antonio Torralba.
\newblock Anticipating visual representations from unlabeled video.
\newblock In \emph{Proceedings of the IEEE conference on computer vision and pattern recognition}, pp.\  98--106, 2016.

\bibitem[Wei et~al.(2022)Wei, Wang, Schuurmans, Bosma, Xia, Chi, Le, Zhou, et~al.]{wei2022chain}
Jason Wei, Xuezhi Wang, Dale Schuurmans, Maarten Bosma, Fei Xia, Ed~Chi, Quoc~V Le, Denny Zhou, et~al.
\newblock Chain-of-thought prompting elicits reasoning in large language models.
\newblock \emph{Advances in neural information processing systems}, 35:\penalty0 24824--24837, 2022.

\bibitem[Welleck et~al.(2019)Welleck, Brantley, Iii, and Cho]{welleck2019non}
Sean Welleck, Kiant{\'e} Brantley, Hal~Daum{\'e} Iii, and Kyunghyun Cho.
\newblock Non-monotonic sequential text generation.
\newblock In \emph{International Conference on Machine Learning}, pp.\  6716--6726. PMLR, 2019.

\bibitem[Williams \& Zipser(1989)Williams and Zipser]{williams1989learning}
Ronald~J Williams and David Zipser.
\newblock A learning algorithm for continually running fully recurrent neural networks.
\newblock \emph{Neural computation}, 1\penalty0 (2):\penalty0 270--280, 1989.

\bibitem[Wu et~al.()Wu, Morris, and Levine]{wulanguage}
Wilson Wu, John~Xavier Morris, and Lionel Levine.
\newblock Do language models plan ahead for future tokens?
\newblock In \emph{First Conference on Language Modeling}.

\bibitem[Yang et~al.(2019)Yang, Dai, Yang, Carbonell, Salakhutdinov, and Le]{yang2019xlnet}
Zhilin Yang, Zihang Dai, Yiming Yang, Jaime Carbonell, Russ~R Salakhutdinov, and Quoc~V Le.
\newblock Xlnet: Generalized autoregressive pretraining for language understanding.
\newblock \emph{Advances in neural information processing systems}, 32, 2019.

\bibitem[Yin et~al.(2024)Yin, Ding, Song, and Zhang]{yin2024semformer}
Yongjing Yin, Junran Ding, Kai Song, and Yue Zhang.
\newblock Semformer: Transformer language models with semantic planning.
\newblock \emph{arXiv preprint arXiv:2409.11143}, 2024.

\end{thebibliography}
\bibliographystyle{colm2025_conference}

\newpage
\appendix
\section{Appendix}
\subsection{Reproducibility statement}
Code and datasets for all experiments are currently in preparation and will be released.

\subsection{Implementation details}\label{appendix:training_details}

\textbf{Training details}: All results are reported on the pretrained-Llama 3.2-1B model. We conducted experiments by sweeping over learning rates of 1e-5, 2e-5, and 1e-6, using the AdamW optimizer with a linear learning rate scheduler for one epoch, and reporting the best result. We use the masked cross-entropy loss specified in \S~\ref{loss_fn}. 
We use $p=0.5$ for all experiments.
All experiments were run on 4xA6000 GPUs or 4xL40S GPUs. 
We will also provide the full list of hyperparameters and release code and datasets used.
\vspace{-1em}
\subsection{Ablations - Autoregressive architectures}
In this section we also compare against other autoregressive architectures. We use mamba as a representative model class for state space models. We observe that using \method{}-Random improves on next token prediction on state space architectures as well.

\begin{table}[ht]
\centering
\small
\begin{tabular}{@{}l@{\hspace{3pt}}l@{\hspace{0pt}}c@{\hspace{5pt}}c@{\hspace{5pt}}c@{\hspace{5pt}}c@{}}
\toprule
       & & \multicolumn{4}{c}{\textbf{Path planning} G(*,*)} \\
       & & G(2,5) & G(5,5) & G(20,5) & G(2,10) \\
\midrule
\multirow{2}{*}{\rotatebox{90}{\footnotesize AR.}} &
      NTP       & \underline{0.50} & \underline{0.20} & \textbf{0.05} & \underline{0.50} \\
      & \method{} &  \textbf{1.0}    &  \textbf{0.998}    &     \underline{0.049}   &   \underline{0.50}    \\
\midrule
\multirow{2}{*}{\rotatebox{90}{\footnotesize Gen.}} &
      NTP       & --  & --  & --  & -- \\
      & \method{} &  \textbf{1.0}    &  \underline{0.997}  &     0.048   &  \textbf{0.511}    \\
\midrule
\multirow{2}{*}{\rotatebox{90}{\footnotesize Spec.}} &
      NTP       & --  & --   & --   & -- \\
      & \method{} &  \textbf{1.0}    &  \textbf{0.998}   &     0.048   &   \underline{0.50}   \\
\bottomrule
\end{tabular}
\caption{Mamba-1.5B - Results on star graph}
\label{tab:mamba_accuracies}
\end{table}
\vspace{-1em}
\subsection{Ablations - Model sizing}\label{appendix:ablations_model_capacity}
To compare the effects of model size on \method{}-Random, we perform on 0.5B (Qwen2.5-0.5B), 1B (Llama-3.2-1B) and 3B (Llama-3.2-3B) models. We do not account for architectural differences between the Qwen 0.5B model and the 1B and 3B Llama models.

The smallest model is unable to solve the longest graph that we test for \(G(2,10)\), while the 1B model is able to solve the graph when allowed to generate \(\rvz\). Finally, the 3B model, is able to solve the graph with only autoregressive generation when trained with \method{}. This hints at \method{} being more effective on larger models, potentially learning better representations, and being easily scalable. Interestingly, larger models can solve the simplest graphs (\(G(2,5), G(5,5)\)) autoregressively. We speculate that this could be due to pre-caching improving with scale as previously observed by \cite{wulanguage}

\begin{table}[ht]
\centering
\small
\begin{minipage}[t]{0.45\textwidth}
\centering
\begin{tabular}{@{}l@{\hspace{3pt}}l@{\hspace{0pt}}c@{\hspace{5pt}}c@{\hspace{5pt}}c@{\hspace{5pt}}c@{}}
\toprule
       & & \multicolumn{4}{c}{\textbf{Path planning} G(*,*)} \\
       & & G(2,5) & G(5,5) & G(20,5) & G(2,10) \\
\midrule
\multirow{2}{*}{\rotatebox{90}{\footnotesize AR.}} &
      NTP       & \underline{0.50} & \underline{0.20} &  0.05  & 0.50 \\
      & \method{} & \textbf{1.0}    & \textbf{1.0}    & \underline{0.874} & \textbf{0.533} \\
\midrule
\multirow{2}{*}{\rotatebox{90}{\footnotesize Gen.}} &
      NTP       & --  & --  & --  & -- \\
      & \method{} & \textbf{1.0}   & \textbf{1.0}   & 0.847 & 0.514 \\
\midrule
\multirow{2}{*}{\rotatebox{90}{\footnotesize Spec.}} &
      NTP       & --  & --  & --  & -- \\
      & \method{} & \textbf{1.0}   & \textbf{1.0}   & \textbf{0.931} & 0.523 \\
\bottomrule
\end{tabular}
\caption{Qwen/Qwen2.5-0.5B}
\label{tab:qwen_accuracies}
\end{minipage}\hfill
\begin{minipage}[t]{0.45\textwidth}
\centering
\begin{tabular}{@{}l@{\hspace{3pt}}l@{\hspace{0pt}}c@{\hspace{5pt}}c@{\hspace{5pt}}c@{\hspace{5pt}}c@{}}
\toprule
       & & \multicolumn{4}{c}{\textbf{Path planning} G(*,*)} \\
       & & G(2,5) & G(5,5) & G(20,5) & G(2,10) \\
\midrule
\multirow{2}{*}{\rotatebox{90}{\footnotesize AR.}} &
      NTP       & \textbf{1.0} & \textbf{1.0} & \underline{0.05} & \underline{0.50} \\
      & \method{} & \textbf{1.0} & \textbf{1.0} & \textbf{1.0}  & \textbf{1.0}  \\
\midrule
\multirow{2}{*}{\rotatebox{90}{\footnotesize Gen.}} &
      NTP       & --  & --  & --  & -- \\
      & \method{} & 1.0  & 1.0  & 1.0  & 1.0 \\
\midrule
\multirow{2}{*}{\rotatebox{90}{\footnotesize Spec.}} &
      NTP       & --  & --  & --  & -- \\
      & \method{} & 1.0  & 1.0  & 1.0  & 1.0 \\
\bottomrule
\end{tabular}
\caption{meta-llama/Llama-3.2-3B}
\label{tab:llama3B_accuracies}
\end{minipage}
\end{table}

\newpage
\section{Star graph}

\subsection {Results - Llama-3.2-1B}

\begin{figure}[ht]
  \centering
  \begin{subfigure}[b]{1.0\textwidth}
    \centering
    \includegraphics[width=\textwidth]{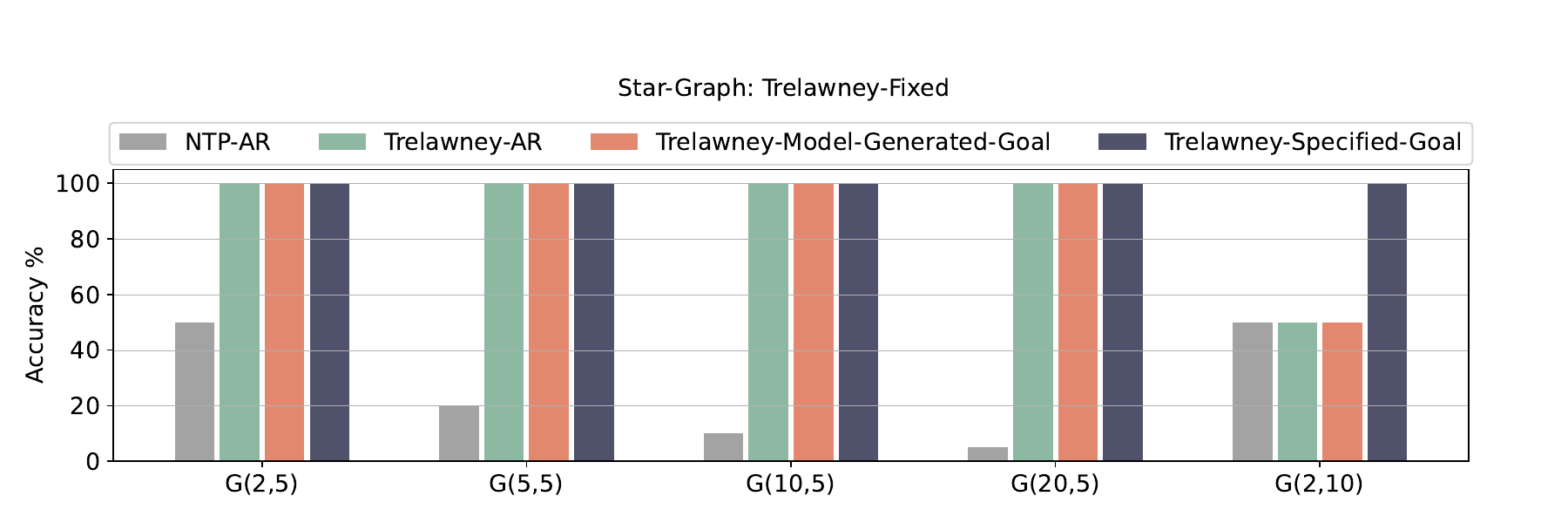}
    \caption{Accuracies \method-{Fixed}}
    \label{fig:accuracies_ours_fixed}
  \end{subfigure}
  
  \vspace{1em}  %
  
  \begin{subfigure}[b]{1.0\textwidth}
    \centering
    \includegraphics[width=\textwidth]{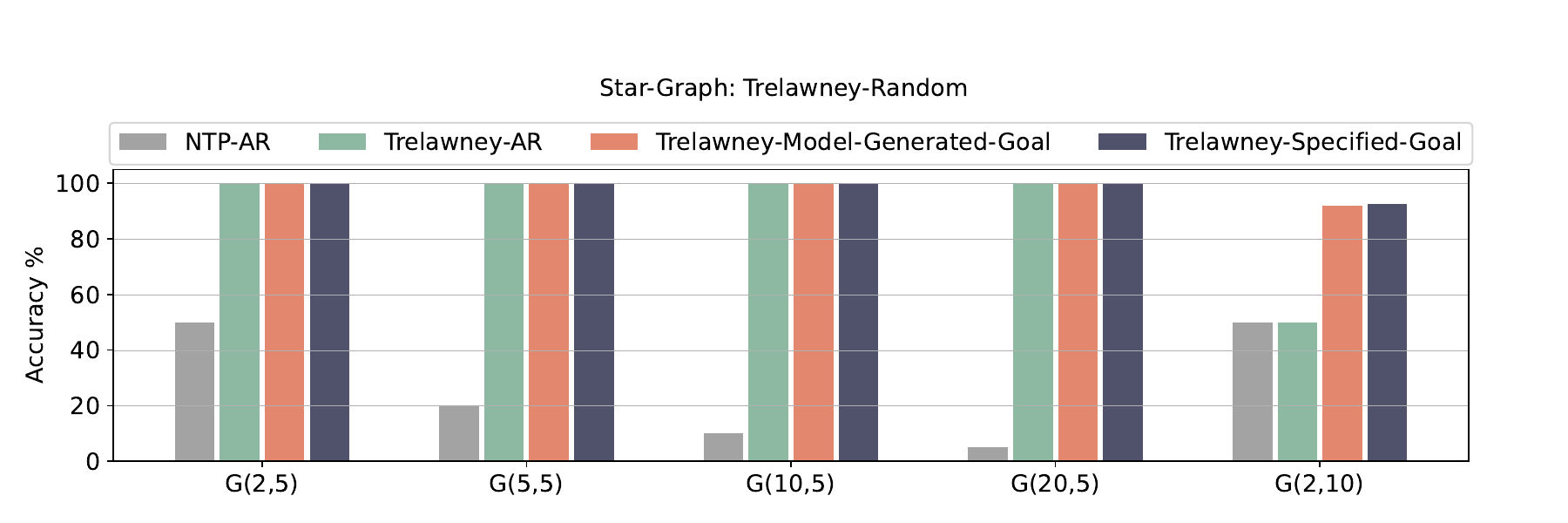}
    \caption{Accuracies \method-{Random}} 
    \label{fig:accuracies_ours_random}
  \end{subfigure}
  
  \caption{Results on Star Graph.}
  \label{fig:accuracies_both}
\end{figure}

\newpage
\section {Algorithmic reasoning}

\subsection{Examples}
We provide examples of the data augmentation schema used in \(D_\text{aug}\) for the strongly connected components task.Looking at the examples in the data, we see many repeated states \(t\) in the trace. The rule used to pick $\rvz$ was likely more beneficial, since the first change in the trace state contains more information than a random trace state.\newline
\label{example:scc}
\begin{tcolorbox}[colback=gray!20, colframe=black, title=Strongly connected components - Examples]
\textbf{Prefix:}
\begin{verbatim}
"strongly_connected_components:
 A: [[0 0 0 0 0 0], 
    [0 0 0 0 0 0], 
    [0 0 0 0 1 0], 
    [0 0 0 1 1 0], 
    [0 0 1 1 1 0], 
    [0 0 0 0 0 1]], 
initial_trace: [0 1 2 3 4 5] trace | scc_id:",

\end{verbatim}
\textbf{Completion:}
\begin{verbatim}
   "[0 1 2 3 4 5], [0 1 2 3 4 5], [0 1 2 3 4 5], [0 1 2 3 4 5], [0 1 2 3 4 5], 
    [0 1 2 3 4 5], [0 1 2 3 4 5], [0 1 2 3 4 5], [0 1 2 3 4 5], [0 1 2 3 4 5],
    [0 1 2 3 4 5], [0 1 2 3 4 5], [0 1 2 3 4 5], [0 1 2 3 4 5], [0 1 2 3 4 5],
    [0 1 2 3 4 5], [0 1 2 3 4 5], [0 1 2 3 4 5], [0 1 2 3 4 5], [0 1 2 3 4 5],
    [0 1 2 3 4 5], [0 1 2 3 4 5], [0 1 2 3 4 5], [0 1 2 3 2 5], [0 1 2 2 2 5],
    [0 1 2 2 2 5], [0 1 2 2 2 5], [0 1 2 2 2 5], [0 1 2 2 2 5], [0 1 2 2 2 5],
    [0 1 2 2 2 5], [0 1 2 2 2 5] | [0 1 2 2 2 5]" 
\end{verbatim}

\textbf{Completion for \method{}--Rule-Based:}
\begin{verbatim}
   "[0 1 2 3 4 5], <T> [0 1 2 3 2 5], </T> [0 1 2 3 4 5], [0 1 2 3 4 5], 
    [0 1 2 3 4 5], [0 1 2 3 4 5], [0 1 2 3 4 5], [0 1 2 3 4 5], [0 1 2 3 4 5], 
    [0 1 2 3 4 5], [0 1 2 3 4 5], [0 1 2 3 4 5], [0 1 2 3 4 5], [0 1 2 3 4 5], 
    [0 1 2 3 4 5], [0 1 2 3 4 5], [0 1 2 3 4 5], [0 1 2 3 4 5], [0 1 2 3 4 5], 
    [0 1 2 3 4 5], [0 1 2 3 4 5], [0 1 2 3 4 5], [0 1 2 3 4 5], [0 1 2 3 4 5], 
    [0 1 2 3 2 5], [0 1 2 2 2 5], [0 1 2 2 2 5], [0 1 2 2 2 5], [0 1 2 2 2 5], 
    [0 1 2 2 2 5], [0 1 2 2 2 5], [0 1 2 2 2 5], [0 1 2 2 2 5] | [0 1 2 2 2 5]" 
\end{verbatim}

\textbf{Completion for \method{}--Random:}
\begin{verbatim}
   "[0 1 2 3 4 5], <T> [0 1 2 3 4 5] </T>  [0 1 2 3 4 5], [0 1 2 3 4 5], [0 1 2 3 4 5], 
    [0 1 2 3 4 5], [0 1 2 3 4 5], [0 1 2 3 4 5], [0 1 2 3 4 5], [0 1 2 3 4 5], 
    [0 1 2 3 4 5], [0 1 2 3 4 5], [0 1 2 3 4 5], [0 1 2 3 4 5], [0 1 2 3 4 5], 
    [0 1 2 3 4 5], [0 1 2 3 4 5], [0 1 2 3 4 5], [0 1 2 3 4 5], [0 1 2 3 4 5], 
    [0 1 2 3 4 5], [0 1 2 3 4 5], [0 1 2 3 4 5], [0 1 2 3 4 5], [0 1 2 3 2 5], 
    [0 1 2 2 2 5], [0 1 2 2 2 5], [0 1 2 2 2 5], [0 1 2 2 2 5], [0 1 2 2 2 5], 
    [0 1 2 2 2 5], [0 1 2 2 2 5], [0 1 2 2 2 5] | [0 1 2 2 2 5]" 
\end{verbatim}

\end{tcolorbox}

\newpage

\subsection {Results - Llama-3.2-1B}
\begin{figure}[ht]
  \centering
  \includegraphics[width=1.0\textwidth]{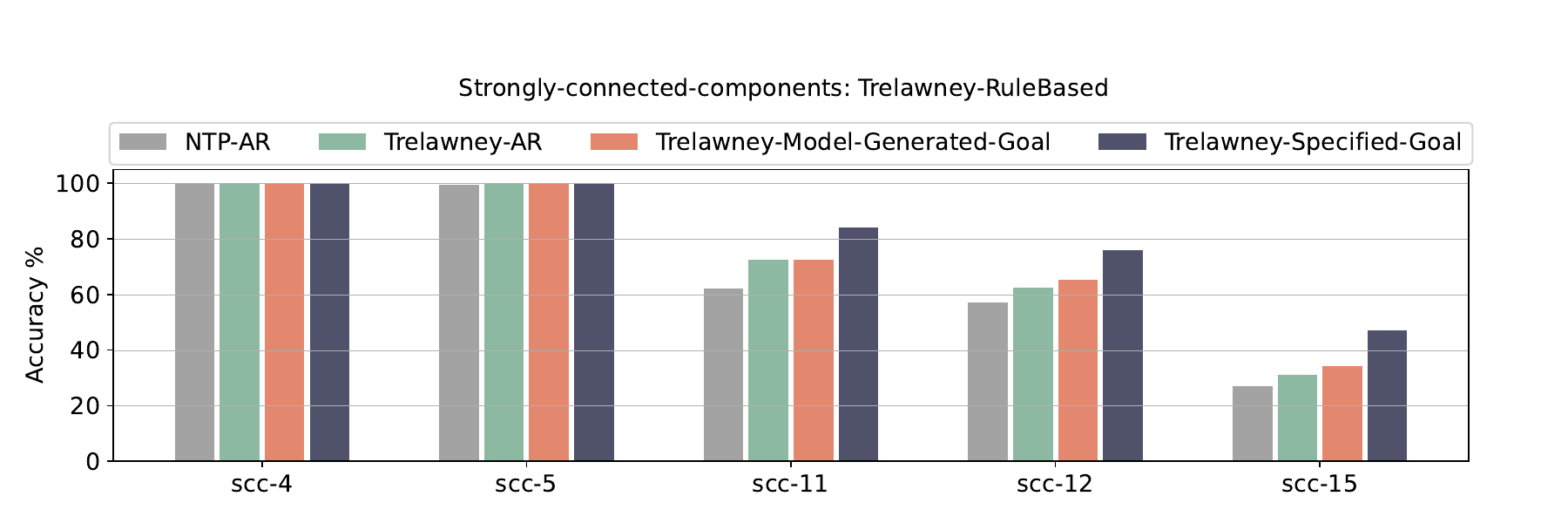}
  \caption{Accuracies - Strongly connected components \ \method-{Rule-Based}}
  \label{fig:accuracies_scc}
\end{figure}
\begin{figure}[ht]
  \centering
  \includegraphics[width=1.0\textwidth]{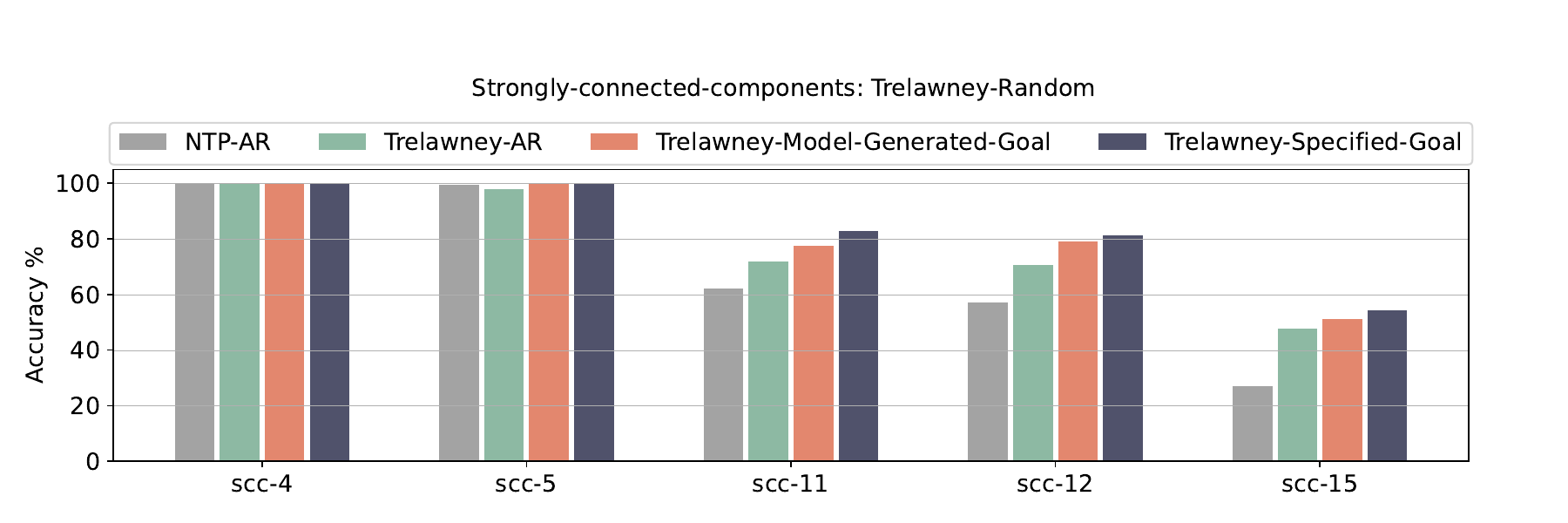}
  \caption{Accuracies - Strongly connected components \ \method-{Random}}
  \label{fig:random_accuracies_scc}
\end{figure}

\newpage
\section{Story generation}
\subsection{Examples}
\begin{figure}[ht]
  \centering
  \includegraphics[trim=1.5cm 16cm 1cm 1cm,clip,width=\textwidth]{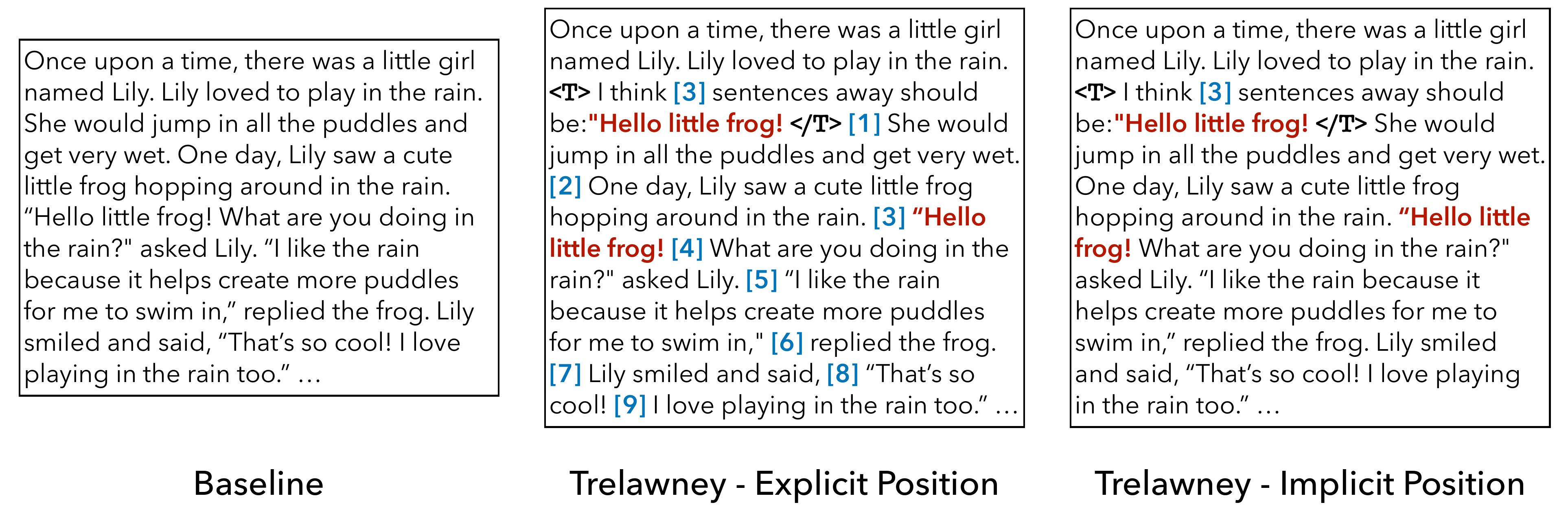}
  \caption{Construction of \(\gD'\) }\textbf{}

  \label{fig:tiny_stories}
\end{figure}

\newpage

\subsection{GPT4-as-Judge}\label{appendix:gpt4_judge}

\paragraph{Unconditional generation.} System prompt used and an example 
\begin{tcolorbox}[colback=gray!20, colframe=black, title=Unconditional generation]
\textbf{System prompt:} You are evaluating two story generation models. First, analyze the outputs and note their properties. Then, make an argument for why one model is better than the other, or state that both models perform similarly. \
                Finally, Respond with WINNER: Story 1/2 (depending on which is better) or respond with DRAW if both are equal.
\newline
\newline
\textbf{User}
\begin{Verbatim}[breaklines=true]
Story 1:
"Once upon a time, there was a naughty bee named Buzzy.Buzzy loved to fly around the big tree and play with the other bees.One day, Buzzy saw a big, juicy flower and wanted to eat it.But the flower was too high up for Buzzy to reach.\n\nBuzzy had an idea!He flew up to the top of the tree and found a long stick.He used the stick to reach the flower and eat it.Buzzy was very happy and flew back down to tell the other bees about his adventure.\n\nThe other bees were impressed with Buzzy's cleverness.They all decided to try and reach the flower too.They worked together and used their wings to fly up to the top of the tree.Buzzy was proud of them and they all shared the delicious flower together.From that day on, Buzzy and the other bees were known as the best bee team in the forest.",  
 
Story 2:
"Once upon a time, there was a naughty bee named Buzzy.Buzzy loved to fly around the big tree and play with the other bees.One day, Buzzy saw a big, juicy flower and wanted to eat it.But the flower was too high up for Buzzy to reach.\n\nBuzzy had an idea!He flew up to the top of the tree and found a long stick.He used the stick to reach the flower and eat it.Buzzy was very happy and flew back down to tell the other bees about his adventure.\n\nThe other bees were impressed with Buzzy's cleverness.They all decided to try reaching the flower too.They worked together and soon they were all able to reach the flower and eat it.Buzzy was proud of his friends and they all had a fun day playing together."    
\end{Verbatim}
\textbf{Judge}
\begin{Verbatim}[breaklines=true]
"Both stories have similar structures, characters, and plotlines. They both follow the adventure of a bee named Buzzy who uses a stick to reach a flower that was too high up. The other bees in both stories are impressed by Buzzy's cleverness and decide to try reaching the flower too. \n\nHowever, there are subtle differences in the way the stories are told. Story 1 ends with a stronger conclusion, stating that Buzzy and the other bees were known as the best bee team in the forest. This gives a sense of accomplishment and closure to the story. On the other hand, Story 2 ends on a more casual note, stating that they all had a fun day playing together. \n\nIn terms of language and style, Story 1 uses more descriptive language and provides more details, such as the bees using their wings to fly up to the top of the tree. Story 2 is more straightforward and uses simpler language.\n\nIn conclusion, while both stories are similar, Story 1 is slightly better due to its stronger conclusion and more descriptive language.\n\nWINNER: Story 1"
\end{Verbatim}

\end{tcolorbox}

\paragraph{Conditional generation}:  We assess if the model was able to reach the specified goal and if the generated stories are coherent. We run two ablations of our method on the baseline trained model 1) Few shot prompts with instructions in natural language - the baseline is unable to reach any of the goals effectively, 2) Few shot prompts (8 examples) with goals specified as tags. Note: we do not use special tokens, only patterns resembling tags. For fairness we report the version with tags in the main paper.

\begin{tcolorbox}[colback=gray!20, colframe=black, title= Conditional generation]
\textbf{System prompt:} You are evaluating two story generation models for goal reaching ability. The goal is specified within \texttt{<LOOKAHEAD>} and  \texttt{</LOOKAHEAD>}, analyze if the stories reach the goal and Then, make an argument for why one model is better than the other, or state that both models perform similarly. \
                Finally, Respond with WINNER: Story 1/2 (depending on which is better) or respond with DRAW if both are equal

\textbf{User}
\begin{Verbatim}[breaklines=true]
Story 1:

"Once upon a time, there was a goat.The goat was outside and it was very hot.<LOOKAHEAD> I think [4] sentences away should be: The goat looked for another way inside but there was none. </LOOKAHEAD>The goat was very thirsty and it wanted to drink some water.\n\nThe goat saw a big tree and it thought it would be a good place to drink water.But when the goat got to the tree, it was too high up and the goat couldn't reach the water.\n\nThe goat looked for another way inside but there was none.So the goat had to stay outside and wait for the water to come down from the sky.\n\nFinally, the water came down and the goat was very happy.It drank the water and felt much better.The goat was very grateful for the water and it thanked the sky for providing it.", 


Story 2:

"Once upon a time, there was a goat.The goat was outside and it was very hot.<LOOKAHEAD> I think [4] sentences away should be: The goat looked for another way inside but there was none. </LOOKAHEAD>Then, the goat saw a big tree and thought it would be a good place to hide.\n\nThe goat went to the tree and climbed up.It was very hot inside the tree, but the goat was happy to be safe.The goat stayed in the tree for a long time, but it was not happy because it was not able to get out.\n\nFinally, the sun went down and the goat was able to get out of the tree.The goat was very tired and hot, but it was happy to be outside again.The goat learned that sometimes it is not good to hide in a tree when it is too hot.",

\end{Verbatim}

\textbf{Judge}
\begin{Verbatim}[breaklines=true]
"Both stories have the same goal: \"The goat looked for another way inside but there was none.\" Story 1 reaches this goal perfectly, as the goat tries to find another way to get water but fails, and the sentence is included verbatim. Story 2, however, does not reach the goal. The goat does look for a place to hide, but it successfully finds one in the tree, and the sentence is not included in the story.\n\nTherefore, Story 1 is better at reaching the specified goal than Story 2. Story 1 not only includes the goal sentence but also builds a coherent narrative around it, while Story 2 deviates from the goal and does not include the sentence.\n\nWINNER: Story 1"
\end{Verbatim}

\end{tcolorbox}

We use GPT-4 as the judge to evaluate stories generated by both models:

{
\renewcommand{\arraystretch}{1.8} 
\begin{table}[ht]
\centering

\begin{tabular}{lccc}
\toprule
\textbf{Story Generation} & \textbf{Baseline} & \textbf{Ours} & \textbf{Draw} \\
\midrule
  Autoregressive                   
    & $0.4526^{+0.494}_{-0.411}$ 
    & $0.4035^{+0.445}_{-0.362}$ 
    & $0.1439^{+0.175}_{-0.116}$   \\
 Few shot natural language v. \method{}                  
    & $0.0000^{+0.006}_{-0.000}$ 
    & $0.8860^{+0.911}_{-0.858}$  
    & $0.1139^{+0.142}_{-0.089}$   \\
 
 Few shot tags v. \method{}               
    & $0.1734^{+0.207}_{-0.144}$      
    & $0.7653^{+0.799}_{-0.729}$      
    & $0.0612^{+0.084}_{-0.043}$   \\

\bottomrule
\end{tabular}
\label{tab:tiny-stories-win-rate}
\caption{Tiny stories win rate with confidence intervals at 95th percentile}
\end{table}
}

\textbf{Failure modes}: Often, both models are unable to reach the goal, then the judge outputs DRAW. In some generations, we note that while the full sentence may not be copied verbatim, we still have coherent generations. In implicit generations, the number of sentences away is less accurate than explicitly specifying them.

\subsection{Perplexity}\label{appendix:perplexity}
WikiText Perplexity on models trained with \method{} are comparable to models trained with standard next token prediction, indicating no noticeable loss in text generation abilities.
{
\renewcommand{\arraystretch}{1.8} 
\begin{table}[ht]
\centering

\begin{tabular}{lccc}
\toprule
\textbf{} & \textbf{Bits-per-byte} ($\downarrow$) & \textbf{Byte-Perplexity} ($\downarrow$) & \textbf{Word-Perplexity} ($\downarrow$)\\
\midrule
  Next-Token-Prediction                   
    & 0.6958
    &  1.6198
    &  13.1865  \\
  \method{}                  
    & 0.6975
    & 1.6217
    & 13.2669  \\
 
\bottomrule
\end{tabular}
\label{tab:wikitext-pplx}
\caption{Perplexity metrics on wikitext}
\end{table}
}

\end{document}